\documentclass[a4paper,conference]{IEEEtran}

\usepackage{booktabs}
\usepackage{amsfonts}

\ifCLASSOPTIONcompsoc
\usepackage[nocompress]{cite}
\else
\usepackage{cite}
\fi
\ifCLASSINFOpdf
\usepackage[pdftex]{graphicx}
\graphicspath{{./pdf/}{./jpeg/}}
\DeclareGraphicsExtensions{.pdf,.jpeg,.png,.jpg}
\else
\usepackage[dvips]{graphicx}
\graphicspath{{eps/}}
\DeclareGraphicsExtensions{.eps}
\fi
\ifCLASSOPTIONcompsoc
\usepackage[caption=false,font=normalsize,labelfont=sf,textfont=sf]{subfig}
\else
\usepackage[caption=false,font=footnotesize]{subfig}
\fi

\usepackage{fixltx2e}

\usepackage{stfloats}

\usepackage{url}
\begin{document}
	
	\title{RectiNet-v2: A stacked network architecture for document image dewarping}
	
	\author{
		\IEEEauthorblockN{Hmrishav Bandyopadhyay}
		\IEEEauthorblockA{Dept. of Electronics and Telecomm. Engg.\\
			Jadavpur University, West Bengal, India\\
			hmrishavbandyopadhyay@gmail.com}
		
		\and 
		
		\IEEEauthorblockN{Tanmoy Dasgupta\IEEEauthorrefmark{1},
			Nibaran Das\IEEEauthorrefmark{2}, and
			Mita Nasipuri\IEEEauthorrefmark{3}}
		\IEEEauthorblockA{Dept. of Computer Science and Engg.\\
			Jadavpur University, West Bengal, India\\
			\IEEEauthorrefmark{1}tdg@ieee.org, \IEEEauthorrefmark{2}nibaran.das@jadavpuruniversity.in and\\ \IEEEauthorrefmark{3}mitanasipuri@gmail.com}
	}

	\maketitle

\begin{abstract}
	With the advent of mobile and hand-held cameras, document images have found their way into almost every domain. Dewarping of these images for the removal of perspective distortions and folds is essential so that they can be understood by document recognition algorithms. For this, we propose an end-to-end CNN architecture that can produce distortion free document images from warped documents it takes as input. We train this model on warped document images simulated synthetically to compensate for lack of enough natural data. Our method is novel in the use of a bifurcated decoder with shared weights to prevent intermingling of grid coordinates, in the use of residual networks in the U-Net skip connections to allow flow of data from different receptive fields in the model, and in the use of a gated network to help the model focus on structure and line level detail of the document image. We evaluate our method on the DocUNet dataset, a benchmark in this domain, and obtain results comparable to state-of-the-art methods.
\end{abstract}


	




\section{Introduction}
Photographing a document with the help of a camera is the most popular method of storing it. With the large scale popularization of mobile devices with inbuilt camera and storage functionalities, capturing document images has been the norm of storing data. These captures, however, are done casually more than often, resulting in distorted and warped images that can be interpreted by humans only, but not by document recognition systems due to large differences in illumination, placement and condition of the documents. For machines to understand data contained in captured document images, dewarping of such images is a necessity.

\par A large number of classical image processing and optimization based methods have been proposed for dewarping document images. These however, fail when curves and folds occur simultaneously in document images, which require a more in-depth and varied  analysis. To rectify these complex document images, deep learning methods have been introduced recently by \cite{bandyopadhyay2020gated}, \cite{ma2018docunet}, \cite{das2019dewarpnet} and \cite{liu2020geometric}. These deep learning methods treat the problem of document dewarping as the prediction of a dense grid that can aid in the dewarping process. The dense grid based approach for dewarping images is preferred to the sparse grid based method as it can effectively capture very fine distortion that a very limited set of dewarping points or a sparse grid cannot. As a result, deep learning methods for document dewarping have been able to dewarp images of a complex nature with significantly higher precision as compared to their image processing counterparts.

\par In our model, similar to \cite{ma2018docunet}, \cite{das2019dewarpnet} and \cite{liu2020geometric}, we use a DNN architecture to predict a dense-grid that can dewarp a document image fed to it. Additionally, we make use of a bifurcated and gated network architecture to predict dense grids from warped document images. More specifically, our contributions can be summed up as :
\begin{itemize}
\item Use of a bifurcated network that takes in images of dimension 256x256 and regresses a dense grid that can unwarp the document represented by the image. This unwarping grid can be interpolated later so that the images are dewarped at their original resolution. The bifurcated network allows us to prevent intermingling of dense-grid values.

\item Use of Residual blocks in the skip connections of the U-Nets used in the stacked module. The use of residual blocks as proposed by \cite{ibtehaz2020multiresunet} enables us to leverage different receptive fields in the skip connections and allows us to pass on information from various levels to the decoder layers.

\item Use of Gated Convolutional Layers in the model architecture, inspired by \cite{GSCNN}. The presence of gates in these layers helps to capture edge and line level data and pass it on in later layers as information which the model has to focus on. In other words, the GCN (Gated Convolutional Network) acts as an attention module to the Secondary U-Net.

\item Use of a Boundary Weighted mean squared loss function that focuses more on the boundary of the dense-grids predicted by a Secondary U-Net. This ensures that poor detection of boundaries by the network is penalized more, and unwarps obtained from the module contain minimal background data of the document image.

\end{itemize}

\section{Previous Works}
In the past several years, we have seen significant progress in the domain of document image dewarping. The methods proposed in past can be summarized briefly into the following categories:
\begin{enumerate}
\item Image Processing based methods 
\item Deep Learning based methods
\end{enumerate}

\subsection{Image Processing Methods}
A plethora of image processing based methods have been proposed and studies have been done on both single and multi image tasks in the field of document dewarping. Efforts have been made to reconstruct 3D views of documents both with the help of additional hardware and with the help of Image Processing Algorithms. 
\par Vision systems have been designed that make use of well calibrated stereo cameras and structured laser light sources to capture a 3 dimensional perspective of the document image. \cite{brown} used a 3D scanning system to create a 3D mesh which is then mapped onto a 2D plane with the help of conformal mapping. \cite{meng2014active} made use of structured laser  beams  for  acquiring shape features from  warped  document images which were later used to generate dewarps of these documents. Further methods utilizing multiple images included \cite{koo2009composition} and \cite{you2017multiview}. While \cite{koo2009composition} matched feature points to register identical areas from images capturing different viewpoints and used that to model the 3D structure, \cite{you2017multiview} recovered 3D point clouds from multiple images and used a modified conformal mapping to perform dewarps.  Although many of these systems are able to procure results of significant quality, their application  is  very severely  bound  due  to  the  limitation  posed  by additional hardware. Multi view system which don't need structured laser beams can do without additional hardware, but still need more than one image, which is hardly available for casually captured documents.

\par A different approach to estimating and generating 3D views of documents by \cite{zhang2009unified} and \cite{tan2005restoring} include extracting shape and structural information from illumination effects in document images. These methods of dewarping images, although free from additional hardware, generally perform very poorly as the estimation step requires primary attention on illumination effect or shading for 3D modeling, which is prone to errors. This makes it highly unsuitable for performing dewarps of natural images.

\par Apart from focusing on reconstructing 3D views of documents, many Image Processing based methods made use of line based features to recognize warps and dewarp images on the basis of that. A coarse to fine technique was proposed by \cite{stamatopoulos2010goal} that performed detection of  word and text lines in coarse scales and used pose normalization in the finer stages to create an unwarp of the document. \cite{ezaki2005dewarping} proposed a global optimization based dewarping technique that recognized and converted warped lines to make them parallel, thus restoring documents from non linearly warped images. \cite{tian2011rectification} simultaneously made use of text line structure and character strokes to estimate a distortion grid and then performed unwarps on document images using this grid.

More sophisticated methods of text line detection involved segmentation and iterative procedures. A segmentation based approach for the detection of warps from text line segments was proposed by \cite{gatos2007segmentation}. Further methods involving segmentation were proposed by \cite{kwon2016method}. Iterative methods demonstrated in \cite{frinken2011novel} and \cite{koo2016text} were significantly better than other image processing methods as they could dewarp images repetitively. These methods made iterative checks while aligning text lines and would perform small dewarps at each iteration, getting an overall better result. The drawback with these iterative techniques was that they were too slow and would take up too much time to dewarp a single document image, making them impractical in most scenarios.
\begin{figure}
\centering
\subfloat[]{	
	\includegraphics[width=35mm]{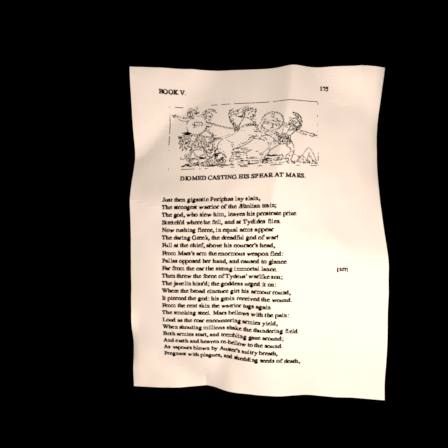}
}\subfloat[]{
	\includegraphics[width=35mm]{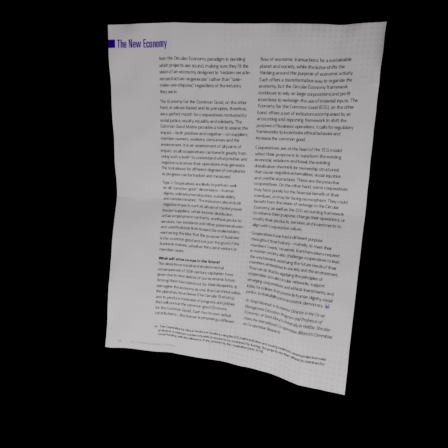}
}\\
\subfloat[]{
	\includegraphics[width=35mm]{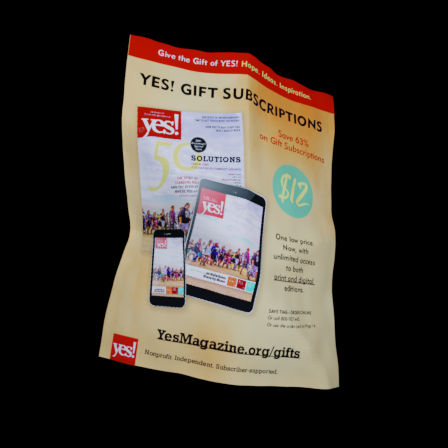}
}\subfloat[]{
	\includegraphics[width=35mm]{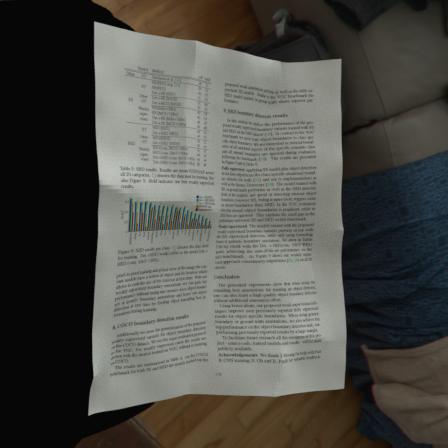}
}\caption{Simulated dataset proposed in \cite{das2019dewarpnet}}
\label{fig:dewarp_data}
\end{figure}


\subsection{Deep Learning based methods}

Application of Deep learning based methods have been relatively low in this domain for quite a long period of time due to the absence of sufficient data for training DNN models. One of the first CNN based method was proposed by \cite{das2017common} where CNNs were used only to detect paper creases for further stages of processing. The first end-to-end CNN model for dewarping document images was proposed by \cite{ma2018docunet} in DocUNet. DocUNet got rid of the requirement of a large scale dataset by synthetically warping scanned images and using them to train the model. The end-to-end network proposed in DocUNet consisted of a stacked U-Net architecture as the backbone. The method of data generation proposed by \cite{ma2018docunet} was used by \cite{liu2020geometric} and \cite{bandyopadhyay2020gated} in their networks.

\par A similar approach of data generation was proposed by \cite{das2019dewarpnet} where focus was given primarily on illumination and shape effects to make the generated data more realistic and to prevent unexpected result when testing on natural images. \cite{das2019dewarpnet} also came up with an end-to-end stacked CNN architecture to dewarp document images that surpassed \cite{ma2018docunet} in dewarping performance by a significant margin.
\section{Dataset}
We make use of the data generation proposed by \cite{das2019dewarpnet} as their data is significantly more realistic and offers better generalization with natural images as compared to \cite{ma2018docunet}. In the dataset proposed by \cite{das2019dewarpnet} 3D shapes and textures of naturally deformed documents were captured and rendered on images with the help of path tracing, taking in many camera positions and a variety of illumination effects and conditions. This allowed the creation of a large scale image dataset with the data being highly realistic, as the illumination and shape effects have been taken from real document images. This not only helps the model to generalize better when used on natural images, but also makes available various forms of ground-truth data including albedo maps, normal maps, depth maps, UV maps, and checkerboard fits, which can be used to further analyze the document structure.
A general representation of the dataset by some of its images is available in Fig. \ref{fig:dewarp_data}.

\section{Methodology}
\subsection{Architecture Overview}
\begin{figure}
\centering
{	
	\includegraphics[width=70mm]{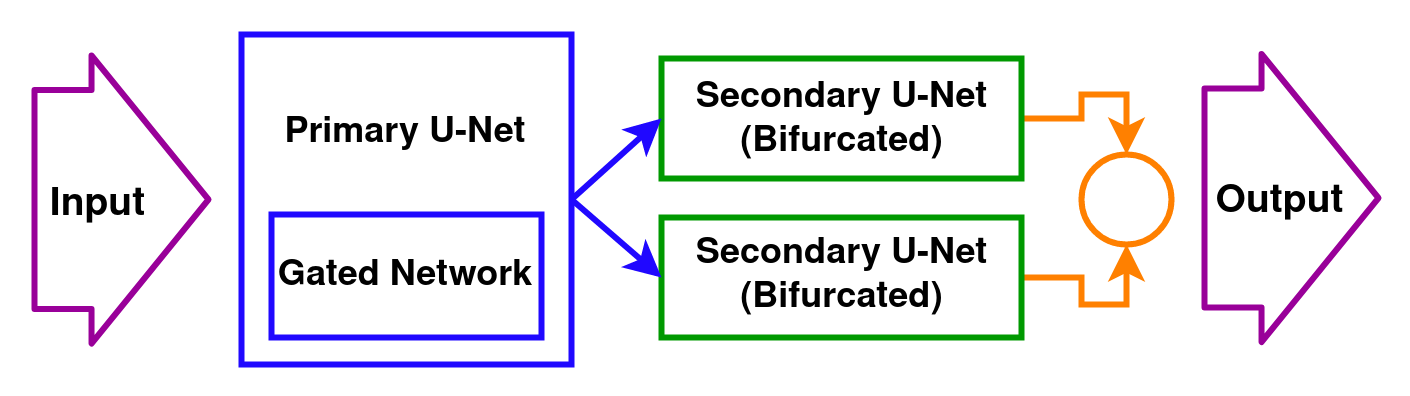}
}
\caption{Complete Architecture}
\label{fig:arch}
\end{figure}

The overall architecture of our method has been expressed in Fig \ref{fig:arch}. We have made significant changes in the stacked U-Net architecture originally proposed by \cite{ma2018docunet}. The major changes lie in the addition of a gated convolutional network for proper processing of line level information and a bifurcation in the secondary U-Net of the stack. Inspired by \cite{ibtehaz2020multiresunet}, we also add residual networks in the skip connections of our model to enhance the features being concatenated in the later stages of our network.

\par The network, as in Fig \ref{fig:arch}, takes in images in the form of batches $D_{n}\in \mathbb{R}^{256\times256\times3}$ and  predicts dense-grids in the form of $D_{n}\in \mathbb{R}^{256\times256\times2}$ where the 2 channels represent the coordinate axes we use for mapping. 

\subsection{Primary U-Net} 
\begin{figure}[!t]
\includegraphics[width=90mm]{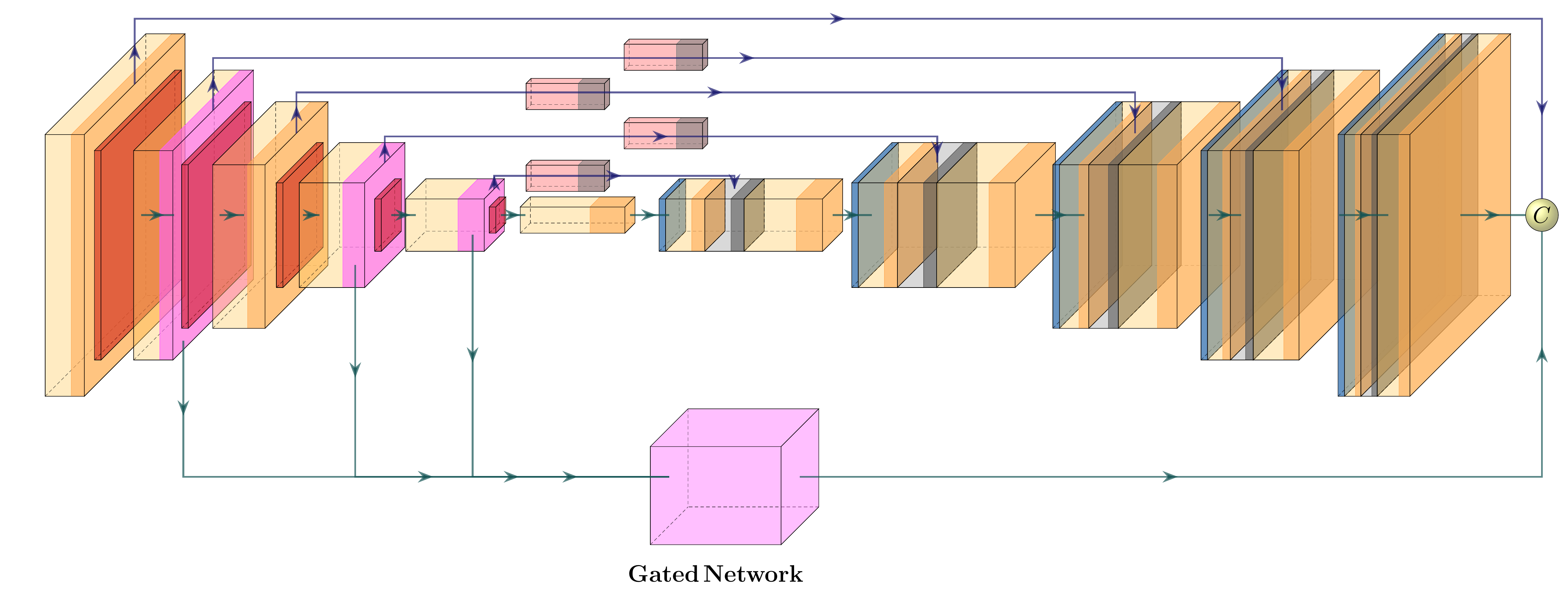}
\caption{Primary U-Net}
\label{fig:arch1}
\end{figure}

The first U-Net of our architecture, aka the primary U-Net is the main block to which we feed the image data. It consists of a series of up-sampling and down-sampling modules with skip connections powered by res-pathways. 

\par The Primary U-Net module as in Fig. \ref{fig:arch1} is fed with images of deformed documents $D_{n}\in \mathbb{R}^{256\times256\times3}$ which gets passed through a series of encoders and to form the bottleneck $B\epsilon\mathbb{R}^{1024\times8\times8}$. The  layers $L_{2}\epsilon\mathbb{R}^{64\times128\times128}$, $L_{4}\epsilon\mathbb{R}^{256\times32\times32}$, $L_{5}\epsilon\mathbb{R}^{512\times16\times16}$ are extracted from these encoders and passed to the gated module $G$. Finally, the decoded outputs $O\epsilon\mathbb{R}^{16\times256\times256}$, gated network outputs $G_{o}\epsilon\mathbb{R}^{16\times256\times256}$ and the initial convolution outputs $X\epsilon\mathbb{R}^{32\times256\times256}$ are  concatenated  to  produce  the  output  of the Primary U-Net $U_{1}\epsilon\mathbb{R}^{64\times256\times256}$.

\begin{enumerate}
\item{\textbf{Residual Path}}: 
\begin{figure}
	\centering
	{	
		\includegraphics[width=60mm]{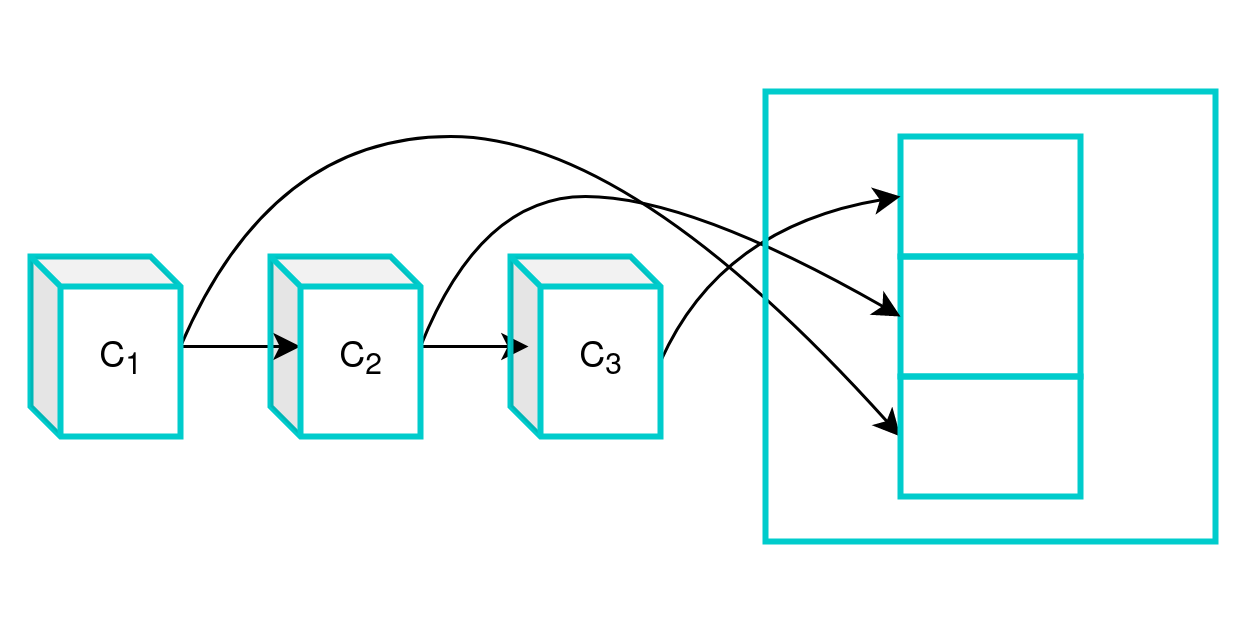}
	}
	\caption{Residual Path}
	\label{fig:arch4}
	
\end{figure}

One of the most ingenious parts of the U-Net architecture is the use of skip-connections. Skip connections in a U-Net help it to retrieve information lost in pooling layers. Inspired by \cite{ibtehaz2020multiresunet}, we make attempts to enhance the information carried by skip connections in a U-Net to the decoder layers by passing it through a Residual Path (res path). The addition of these pathways as shown in Fig. \ref{fig:arch4}, enable the decoder to get more spatial information, which allows the network to deal with images at various scales. The three layers of the res path have convolutions that have a receptive field of $3\times3$, $5\times5$,  and $7\times7$ respectively. The concatenation of data processed through various receptive fields enables us to pass on more spatial information to the decoder than that would have been possible otherwise.
\item{\textbf{Gated Convolutional Network}}: 

\begin{figure}
	\centering
	{	
		\includegraphics[width=70mm]{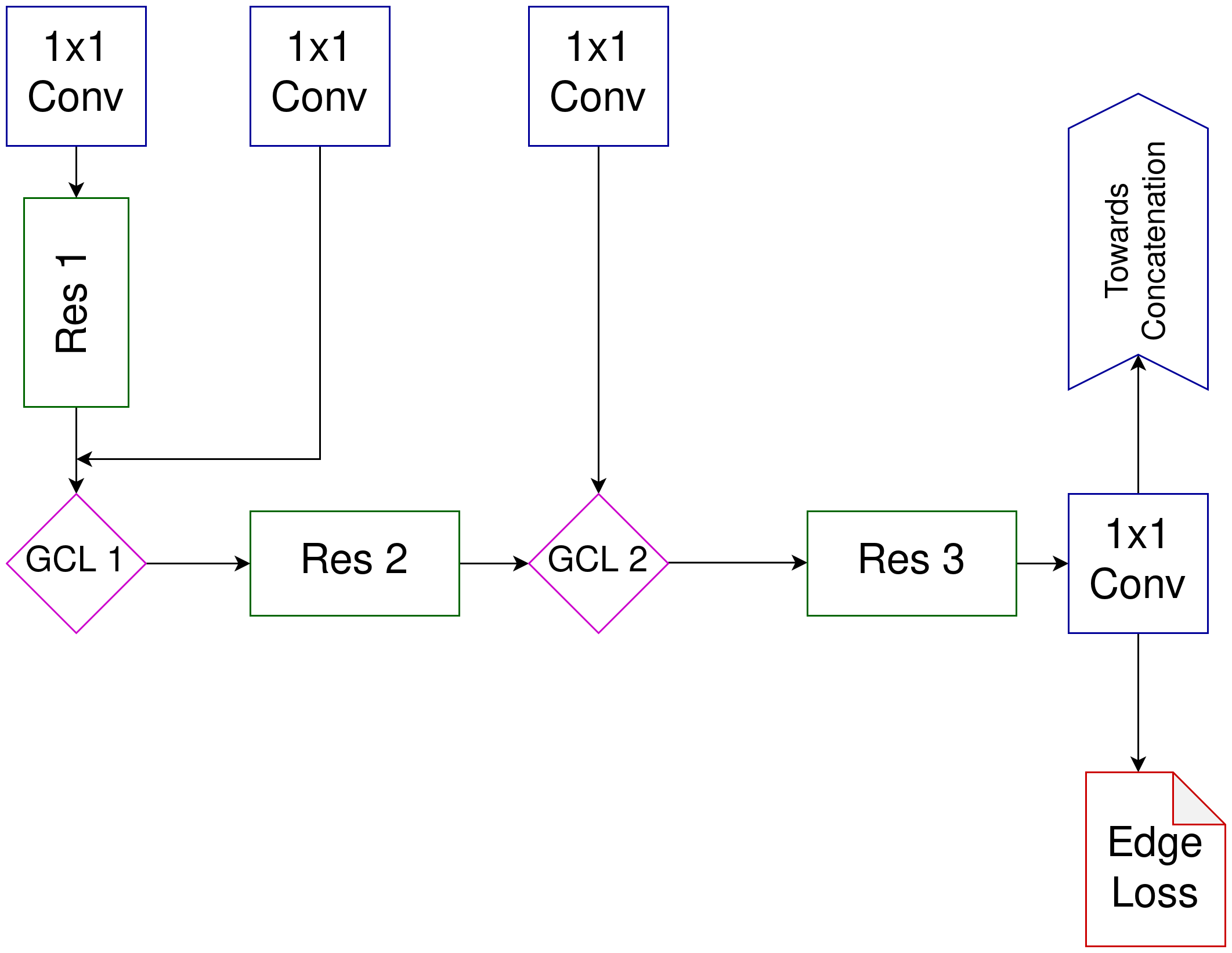}
	}
	\caption{Gated Convolutional Network}
	\label{fig:arch3}
	
\end{figure}
The Gated Network as in Fig. \ref{fig:arch3} works on data extracted in the layers before the 2nd, 4th and 5th poolings. It is built up from Gated Convolutional Layers (GCLs) as proposed in \cite{GSCNN}.The presence of gates in the network architecture helps us to segregate line level data from the image, which is later fed as attention information to the Secondary U-Net. This helps the secondary U-Net to focus more on textual regions and perform better dewarps of document images. 

\end{enumerate}

We train the Primary U-Net along with the Gated Network on the edge loss which is calculated as the binary cross entropy between the output from the GCN and a canny edge filter of the input. The mathematical form of the edge loss can be expressed as :
\begin{equation}
\mathcal{L}_{edge}=-\frac{1}{N}\sum_{i=0}^{N}y_i.log(\hat{y_{i}})+(1-y_{i}).log(1-\hat{y_{i}})
\end{equation}
Where $\hat{y_{i}}$ represents pixel wise value of the predicted output and $y_{i}$ gives the ground truth measure.
\subsection{Secondary U-Net}
\begin{figure}[!h]
\centering
{	
	\includegraphics[width=\columnwidth]{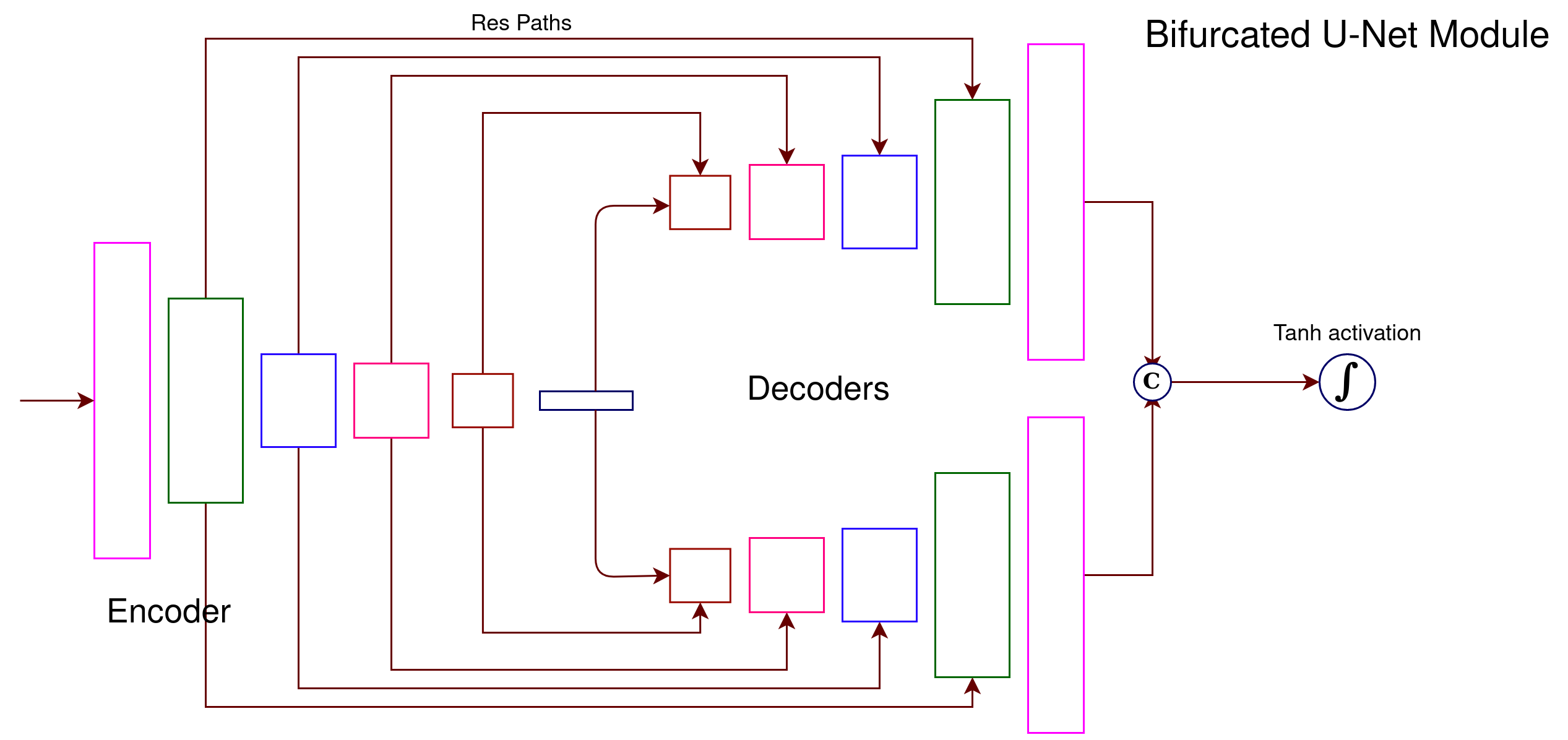}
}
\caption{Secondary U-Net}
\label{fig:arch5}
\end{figure}

Convolutional networks fail badly when dealing with coordinate data as shown by \cite{coordconv}. To deal with this issue better, \cite{das2019dewarpnet} makes use of the coord-conv module as suggested in \cite{coordconv}. However, we find that a more task specific network can be designed which can enhance the ability of CNNs while working with dense grid predictions of 2D document images.

\par The general CNN works by summing up computed data across all input channels for specific window sizes. Ultimately the number of channels in the output is the number of filters that the convolutional block contains. During the summation, the CNN merges data from multiple channels together into vectors and then uses the merged data in the later stages to predict Dense-grids from images. Dense-grids, which are expressed as a set of 2D arrays containing coordinate points from the X and the Y axes of the image, however, don't have much inter-relationship at the channel level and mixing up of the channels at each stage of the network, like a single decoder would have done, doesn't work well. In other words, using a single decoder in the final U-Net block would mean that although	information is extracted in all blocks, it is merged together at each layer and only the last two convolutional filters would be responsible to decode or separate the grid values into their respective channels for the final output. To get over this issue, we have come up with the usage of multiple decoder blocks for the single secondary U-Net encoder, so that channels in the dense grid output are developed separately.


\par The output of the Primary U-Net $U_{1}\in \mathbb{R} ^{64\times256\times256}$ is fed as input to the secondary U-Net as in Fig. \ref{fig:arch5}. The Bottleneck $B\in \mathbb{R} ^{1024\times8\times8}$ is split into $B_{1}\in \mathbb{R} ^{512\times8\times8}$ and $B_{2}\in \mathbb{R} ^{512\times8\times8}$. These blocks go through the decoders to give outputs $O_{1}\in \mathbb{R} ^{256\times256\times1}$ and $O_{2}\in \mathbb{R} ^{256\times256\times1}$, which are concatenated and normalized by a Tanh activation function to get the final grid $g\in \mathbb{R} ^{256\times256\times2}$.
\begin{figure*}
	\centering
	\subfloat{	
		\includegraphics[width=25mm,height=30mm]{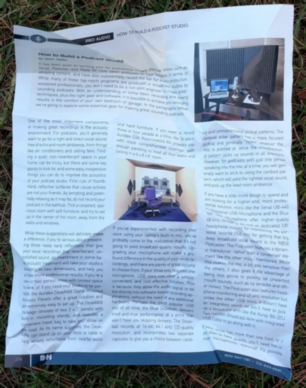}
	}\hspace*{-0.5em}\subfloat{
		\includegraphics[width=25mm,height=30mm]{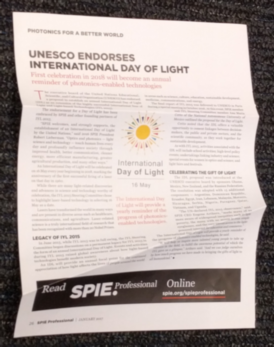}
	}\hspace*{-0.5em}\subfloat{
		\includegraphics[width=25mm,height=30mm]{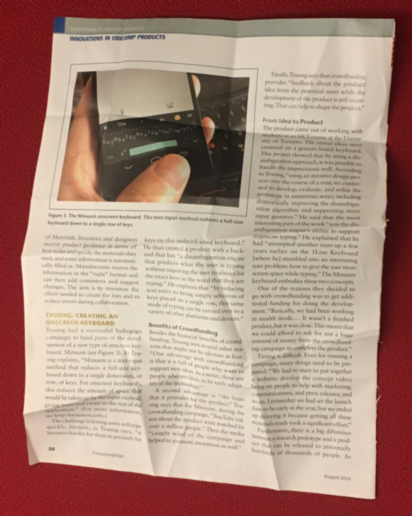}
	}\hspace*{-0.5em}\subfloat{
		\includegraphics[width=25mm,height=30mm]{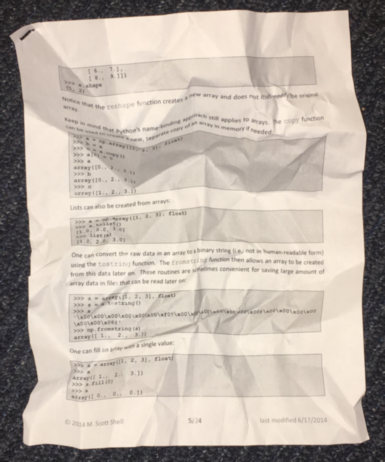}
	}
	\hspace*{-0.5em}\subfloat{
	\includegraphics[width=25mm,height=30mm]{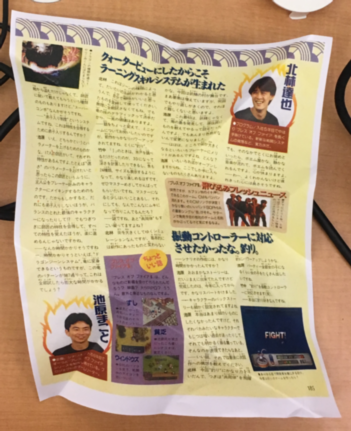}
	}
\hspace*{-0.5em}\subfloat{
	\includegraphics[width=25mm,height=30mm]{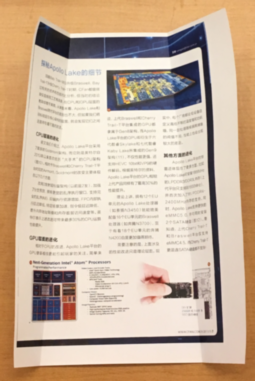}
}
	\\[-0.7em]
	\subfloat{	
		\includegraphics[width=25mm,height=30mm]{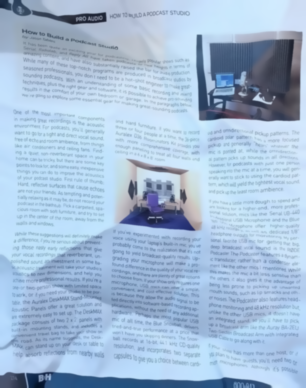}
	}\hspace*{-0.5em}\subfloat{
		\includegraphics[width=25mm,height=30mm]{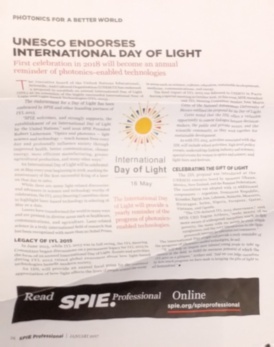}
	}\hspace*{-0.5em}\subfloat{
		\includegraphics[width=25mm,height=30mm]{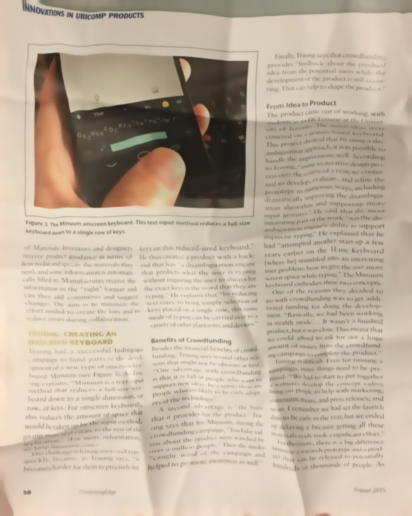}
	}\hspace*{-0.5em}\subfloat{
		\includegraphics[width=25mm,height=30mm]{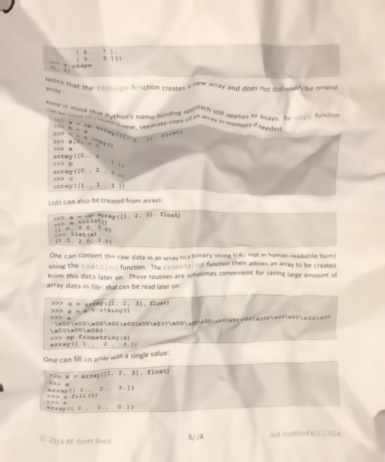}
	}
	\hspace*{-0.5em}\subfloat{
	\includegraphics[width=25mm,height=30mm]{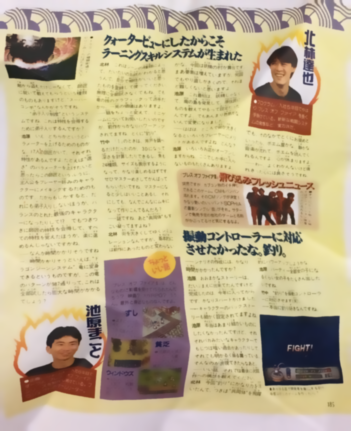}
	}
\hspace*{-0.5em}\subfloat{
	\includegraphics[width=25mm,height=30mm]{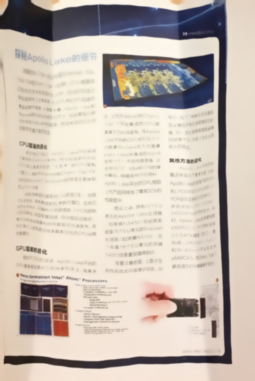}
}
\\[-0.7em]
\subfloat{	
	\includegraphics[width=25mm,height=30mm]{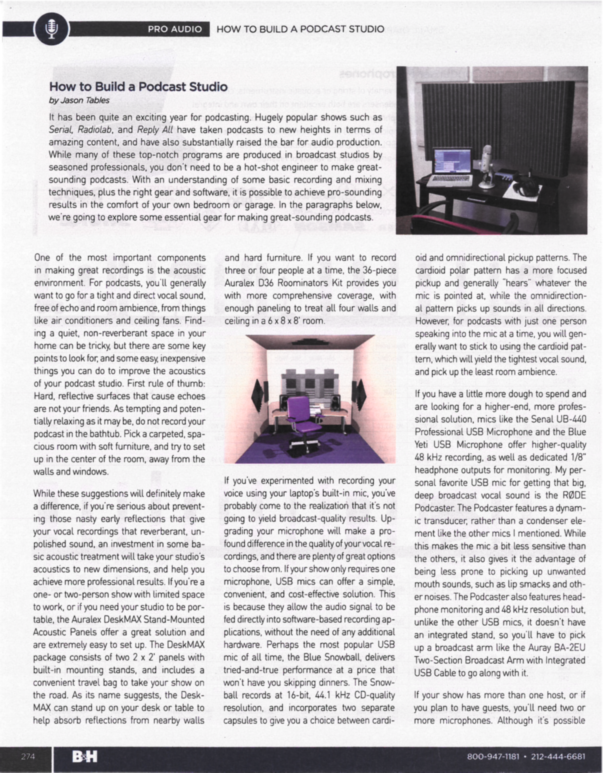}
}\hspace*{-0.5em}\subfloat{
	\includegraphics[width=25mm,height=30mm]{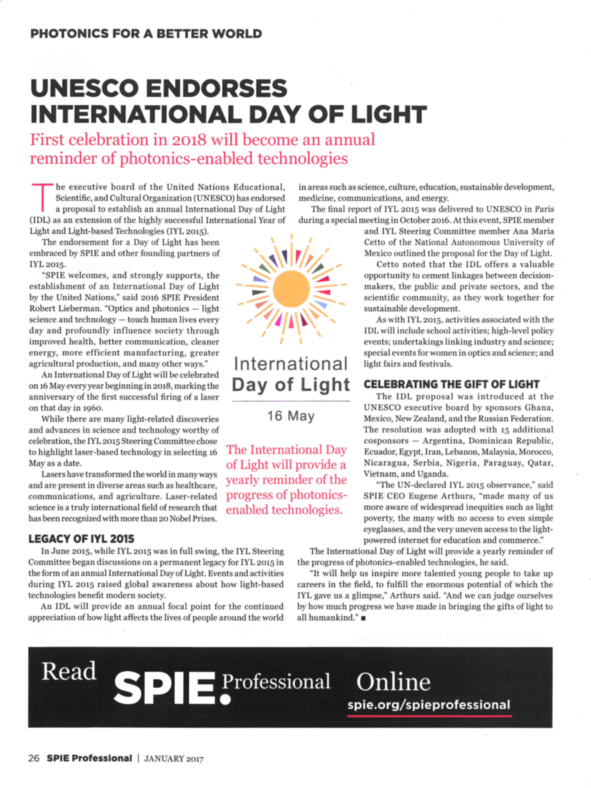}
}\hspace*{-0.5em}\subfloat{
	\includegraphics[width=25mm,height=30mm]{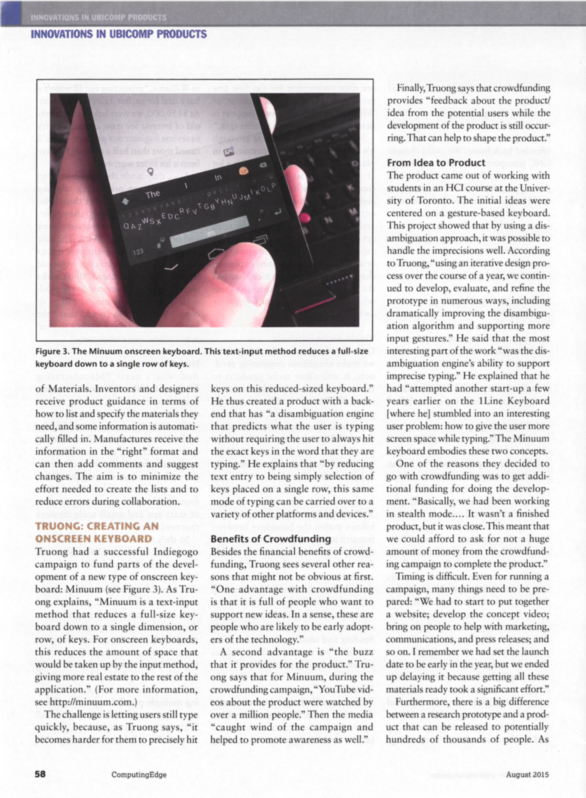}
}\hspace*{-0.5em}\subfloat{
	\includegraphics[width=25mm,height=30mm]{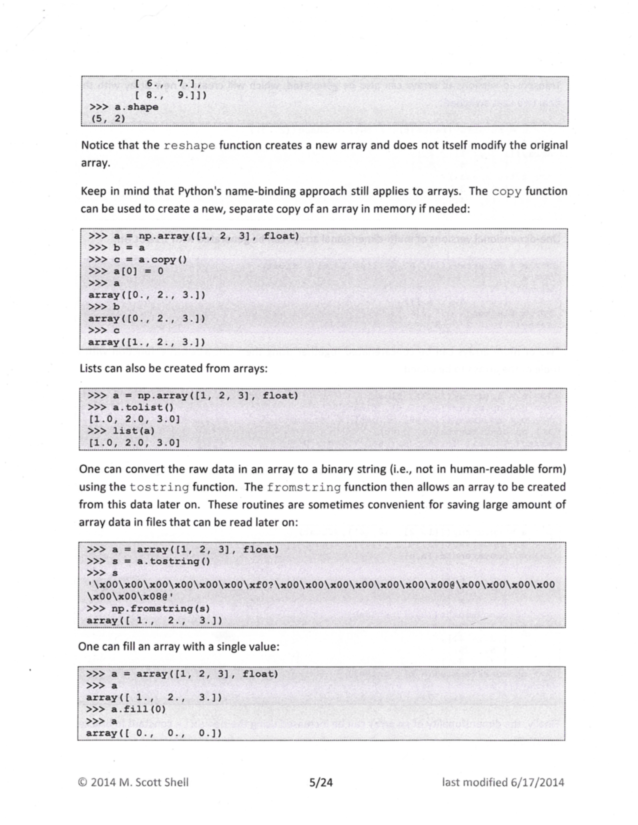}
}
\hspace*{-0.5em}\subfloat{
	\includegraphics[width=25mm,height=30mm]{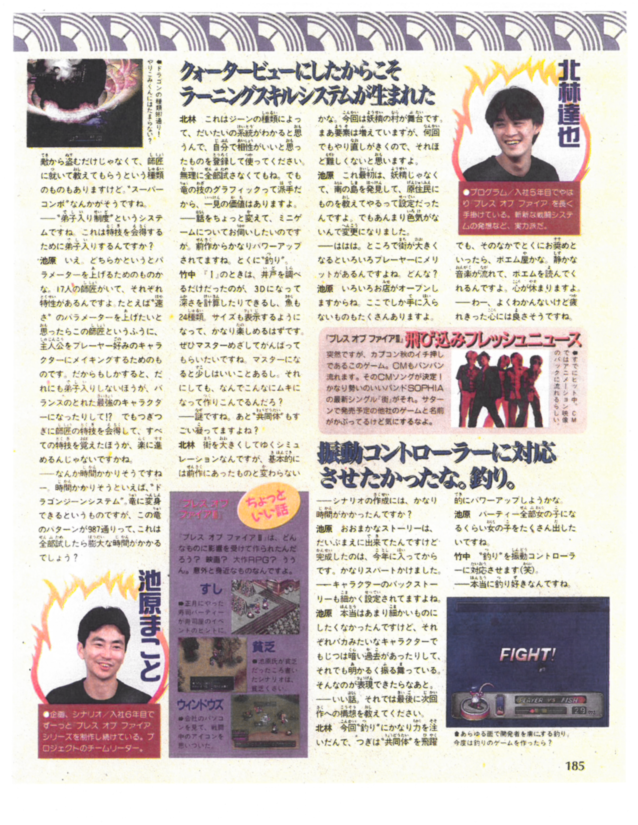}
	
}
\hspace*{-0.5em}\subfloat{
	\includegraphics[width=25mm,height=30mm]{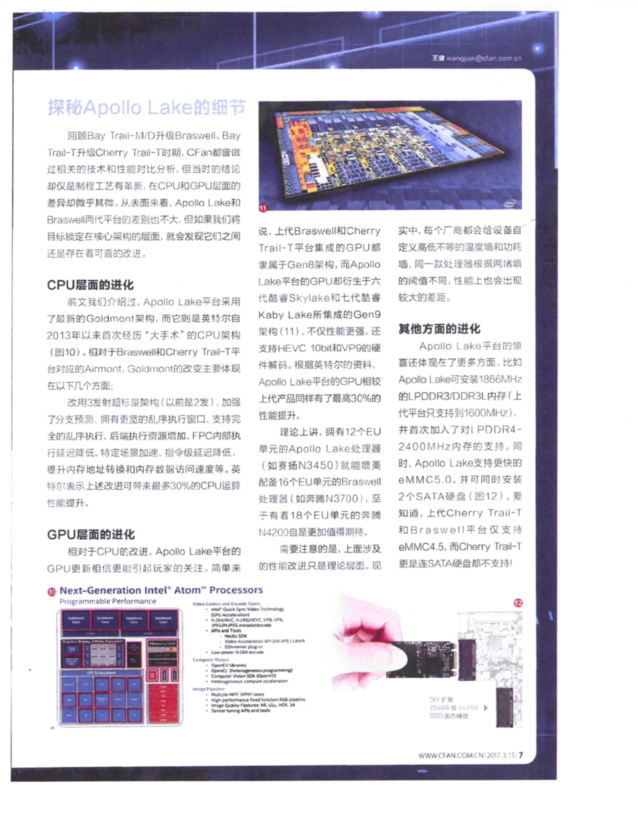}
}
	\caption{Rows from top to bottom: Cropped Images; Results with our methods; Ground Truth}
	\label{fig:results}
\end{figure*}

\par The Secondary U-Net is trained on the grid loss, a weighted Mean Squared Loss between the dense-grid predicted by the network and the ground truth. The weights of the mean squared loss are built in such a way that a wrong prediction of the boundary and the immediate surroundings of the boundary is penalized more as compared to possible errors in the interior. This is done with an intent of making the network focus on the boundaries of documents in the images it works with. The grid loss is used to train the entire network while the edge loss acts as a supportive loss specifically for the Primary U-Net. Mathematically, we can express the grid loss as:

\begin{equation}
\mathcal{L}_{grid}=W\cdot\big[\frac{1}{N}\sum_{i=0}^{N}(g_{i}-\hat{g_{i}})^2\big]\end{equation}
with $W$ being the boundary-weight in the loss function.

The combined loss function can be expressed as:
\begin{equation}
\mathcal{L}=\mathcal{L}_{grid}(S(P(z_{i})),\hat{g}_{i})+\lambda\mathcal{L}_{edge}(G(z_{i}),\hat{y}_{i})
\end{equation}
Where the Gated Convolutional Network is expressed as $G$ and the Primary and the Secondary U-Nets are expressed as $P$ and $S$ respectively.
We take the value of $\lambda$ as 0.9 for all our experiments.

\subsection{Post-Processing}
We further post-process the dewarped document images as part of our dewarping module with the help of bilateral filtering which smoothens folds and crumples. As a variation to post-processing by \cite{bandyopadhyay2020gated} which leaves textual regions blurred in some images, we maintain the level of filtering used by monitoring the blur before and after the filter by computing the variance of a Laplacian filter on the image.
\begin{table}[h]
	\centering 
	\caption{MS-SSIM and LD values on comparison with other methods\label{tab:techniques}}

	\begin{tabular*}{\linewidth}{p{140pt}p{50pt}p{20pt}}    \toprule
		\centering
		\textbf{Method} & \textbf{MS-SSIM} $\uparrow$ &\textbf{LD}$\downarrow$\\\midrule
		 \centering
		 \cite{tian2011rectification}&0.13 & 33.69\\
		 \centering
		 DocUNet, \cite{ma2018docunet}&0.4103&14.08\\
		 \centering
		 AGUN, \cite{liu2020geometric} &0.4491&12.06\\
		 \centering
		 Output from Network&0.4437&10.71\\
		 \centering
		 \textbf{Post-Processed Output} &\textbf{0.4500}&\textbf{10.40}\\
		 \centering
		 DewarpNet, \cite{das2019dewarpnet}&0.4658&8.98\\
		\hline
	\label{table:tab2}
	\end{tabular*}
\end{table}

\section{Experiments}

For comparison of the dewarp quality of our methods, metrics like MS-SSIM (Multi Scale Structural Similarity Index), SSIM (Structural Similarity Index), and LD(Local Distortion) are used. Following \cite{ma2018docunet} and \cite{das2019dewarpnet}, all our results have been presented on the DocUNet benchmark. For the calculation of MS-SSIM and LD, our images have been scaled to approximately 598,400 sq. pixel areas (880x680), while we use original images for the calculation of SSIM along the levels to gauge the quality of dewarp at various levels in Table \ref{table:tab1}. The MS-SSIM values for our methods at the original resolution and comparison with RectiNet by \cite{bandyopadhyay2020gated} can be found in the supplementary. We tabulate the results of our experiments on the benchmark both before and after the  post-processing step in Table \ref{table:tab2}. A subset of the outcomes of our method is displayed in Fig. \ref{fig:results} along with the input and corresponding scanned ground truth. 

\par From Table \ref{table:tab1} and Table \ref{table:tab2}, we infer that our method outperforms methods proposed by \cite{bandyopadhyay2020gated}, \cite{ma2018docunet} and \cite{liu2020geometric} in MS-SSIM while outperforming the method proposed by \cite{das2019dewarpnet} in SSIM measured at original resolution.

%


\begin{table}[h]
	\centering 
	\caption{SSIM values on varying levels}
	\begin{tabular*}{\linewidth}{p{40pt}p{60pt}p{50pt}p{70pt}}\toprule  
		
		\textbf{Level} &\textbf{Our Method} &\textbf{RectiNet}&\textbf{DewarpNet}\\\midrule	 \centering\textbf{Original} &\textbf{0.507434}&\textbf{0.548736}& \textbf{0.493146}\\
		\centering2&0.455497&0.482410&0.445168\\
		\centering3&0.405827&0.409842&0.402247\\
		\centering4&0.391321&0.352511&0.402949\\
		\centering5&0.501857&0.347734&0.544776\\
		\centering6&0.600473&0.465316&0.644614\\
		\centering7&0.649300&0.556033&0.673603\\

		\hline
		\label{table:tab1}
	\end{tabular*}
\end{table}


\section{Conclusion}

In this paper, we have proposed a method to dewarp document images by recognizing their structure and predicting dense-grids for mappings. We further demonstrate the effectiveness of a  bifurcation of the traditional U-Net and the addition of a gated module and residual pathways by comparing our method with state-of-the-art methods on the DocUNet benchmark.

\par However, we find that the performance of our method is not satisfactory in certain respects. In-spite of carefully set thresholds in the post-processing setup proposed by us, we observe a few images get blurred after post-processing due to contrast issues. We further observe the architecture proposed by us fails to dewarp document images which are not cropped to exact fit, demonstrating the necessity of a localization procedure for the same.

\par We find that MS-SSIM as a metric does not provide as much attention to line level detail as it does to overall image structure, texture etc. The area dependency of MS-SSIM and LD also causes them to give highly varied results for the same distortion level in images of different areas. Thus, future work on an area independent  standardized metric is highly necessary for proper evaluation of results.

\section*{Acknowledgments}
All of the experiments demonstrated in this paper have been carried out in the Center for Microprocessor Application for Training Education and Research (CMATER), Jadavpur University on hardware infrastructure provided by Science and Engineering Research Board (SERB), India (Ref.\# SB/S3/EECE/054/2016)

\bibliography{main}

\end{document}